\documentclass[10pt,twocolumn,letterpaper]{article}

\usepackage[pagenumbers]{cvpr} 

\usepackage{graphicx}
\usepackage{amsmath}
\usepackage{amssymb}
\usepackage{booktabs}
\usepackage{kotex}
\usepackage{comment}
\usepackage{booktabs}
\usepackage{multirow}
\usepackage{threeparttable}
\usepackage{tabularx} 
\usepackage{amssymb}
\usepackage{pifont}
\usepackage{algorithm} 
\usepackage{algpseudocode}

\newcommand{\cmark}{\text{\ding{51}}}
\newcommand{\xmark}{\text{\ding{55}}}

\usepackage[pagebackref,breaklinks,colorlinks]{hyperref}

\usepackage[capitalize]{cleveref}
\crefname{section}{Sec.}{Secs.}
\Crefname{section}{Section}{Sections}
\Crefname{table}{Table}{Tables}
\crefname{table}{Tab.}{Tabs.}

\def\cvprPaperID{*****} 
\def\confName{GSAI}
\def\confYear{2023}

\begin{document}

\title{NeuroFlow: Development of lightweight and efficient model integration scheduling strategy for autonomous driving system}  

\author{
Eunbin Seo\\
Seoul National University\\
{\tt\small rlocong339@snu.ac.kr}
\and
Gwanjun Shin\\
Seoul National University\\
{\tt\small shinkansan@snu.ac.kr}
\and
Eunho Lee\\
Seoul National University\\
{\tt\small eunho1124@snu.ac.kr}
}

\maketitle

\begin{abstract}
    This paper proposes a specialized autonomous driving system that takes into account the unique constraints and characteristics of automotive systems, aiming for innovative advancements in autonomous driving technology. The proposed system systematically analyzes the intricate data flow in autonomous driving and provides functionality to dynamically adjust various factors that influence deep learning models. Additionally, for algorithms that do not rely on deep learning models, the system analyzes the flow to determine resource allocation priorities. In essence, the system optimizes data flow and schedules efficiently to ensure real-time performance and safety. The proposed system was implemented in actual autonomous vehicles and experimentally validated across various driving scenarios. The experimental results provide evidence of the system's stable inference and effective control of autonomous vehicles, marking a significant turning point in the development of autonomous driving systems.
\end{abstract}

\section{Introduction}
\label{sec:intro}
Autonomous driving technology is one of the groundbreaking areas leading the innovation in the automotive industry, enabling vehicles to operate, perceive the environment, and reach destinations safely without driver intervention. The Society of Automotive Engineers(SAE) has established a classification system ranging from Level 0 to Level 5 to indicate the autonomy levels of autonomous vehicles. These levels are defined based on the extent of driver involvement, where Level 0 requires the driver to have complete control, and Level 5 represents fully autonomous driving without any need for driver intervention\cite{SAEJ3016}. Particularly, Level 3, positioned in the intermediate stage of autonomous driving technology, allows the vehicle to manage specific tasks and scenarios without driver intervention. This level is anticipated to provide benefits such as enhanced convenience, increased safety, reduced driver fatigue, improved accessibility and mobility\cite{Caradas}.

Driven by these expectations, research in the fields of perception, positioning, judgment, and control has been at the core of developing various models and algorithms for autonomous driving technology. However, research related to AI systems for autonomous driving is still in its nascent stages. Autonomous driving systems have predominantly been developed on the Robot Operating System (ROS), an open-source robot software platform that offers a variety of tools for the design, control, visualization, and simulation of robot systems \cite{quigley2009ros}. Autoware provides an open-source solution by developing recognition, judgment, and operation based on inter-process communication, bag files, and launch files \cite{autoware}. Apollo enhances system safety and performance by incorporating ROS decentralization features, shared memory, and protobuf functionalities \cite{apolloplatform}. D3 introduces a dynamic deadline-driven execution model beyond ROS and proposes ERDOS for its implementation \cite{10.1145/3492321.3519576}.

Creating autonomous driving systems requires careful consideration of the unique constraints and characteristics specific to these systems. 1) There are limitations concerning power consumption, heat generation, and size due to the need to integrate the system into vehicles. 2) Deep learning models must operate across various platforms, and programs with substantial CPU and RAM loads, such as raw data pre-processing of sensors, need to run simultaneously. The presence of program dependencies results in the coexistence of asynchronous and synchronous characteristics of unrelated flow. 3) Real-time performance and safety assurance are paramount. Certain deep learning models exhibit dynamic runtime changes based on input data, making it essential to consider input size.

Understanding driving context and managing computing resources, memory allocation, and overall program flow are crucial to ensuring accurate results from each model and algorithm. For instance, in scenarios like intersections, expanding the batch size for detecting a broader range of target vehicles or adjusting the speed to ensure perception results within a valid time limit on highways are necessary. Additionally, predicting runtime for CPU-intensive and GPU jobs allows for considerations in multiprocessing and distributed inference.

In this context, this paper focuses on the constraints and characteristics of autonomous driving technology, proposing a specialized lightweight and scheduling strategy tailored to autonomous driving systems, referred to as DynaNode. This approach analyzes the data flow within the autonomous driving system in the form of a graph. For programs utilizing deep learning models, it adjusts runtime platforms and input batches. For programs without deep learning models, it determines weights for ancestor nodes and end nodes to adjust the parameters of scheduler. Ultimately, the proposed system was implemented in actual vehicles, validating stable inference from models and the actual driving performance of the vehicles.

The key contributions of our research can be summarized as follows:
\begin{itemize}
    \item We comprehend the constraints and characteristics of autonomous driving systems, enabling us to understand driving context and control factors influencing deep learning models.
    \item We enhance the system's real-time performance and safety through the analysis of data flow and the dynamic adjustment of weights, ensuring optimal system operation.
    \item We validate the proposed system by integrating it into actual vehicles, verifying stable inference from models and the vehicles' real-world driving performance.
\end{itemize}


\section{Related Works}

\subsection{Runtime Prediction in Deep-Learning models}
Runtime prediction in DNN models is influenced by the architecture of the model. In real-time critical systems such as autonomous driving, CNN models are predominantly utilized and these models's runtime prediction is influenced model, operator, kernel, and platform. In the field of Deep Neural Networks (DNNs), core operators like Conv2D, ReLU, and BatchNorm are standardized, despite varied model architectures. These operators, with their specific parameters, significantly affect inference latency and the overall efficiency of DNN models. However, model-level predictions often struggle with accuracy, especially with unfamiliar model graphs. This challenge is addressed by nn-Meter \cite{zhang2021nn}, which recognizes the immense prediction space created by the potential connections in a model's Directed Acyclic Graph (DAG). Given these limitations, finer-grained approaches, such as operator-level predictions, are more effective, yet they can miss critical graph optimizations in edge devices. To overcome these shortcomings, recent research has shifted towards kernel-level prediction. This advanced approach provides more detailed and optimization-aware predictions, leading to improved accuracy and reliability in forecasting model performance across different hardware platforms. Furthermore, attention-based models, which have gained popularity as much as CNN models in recent times, typically require higher computational operations. However, given the diversity of Deep Neural Network (DNN) models and platforms in our research (e.g., IPC, Jetson Xavier Orin), we consider both kernel-level, model-level and rapid attention-based model prediction in our analysis. This comprehensive approach enables us to effectively address the complexities and variabilities inherent in different DNN architectures and hardware environments.

\subsection{Autonomous Driving System}
Autonomous vehicles are equipped with an array of sensors (e.g., Cameras, LiDAR, Radar, GPS) and are powered by ECUs or computational units. Then, Autonomous Driving Systems usually comprise a pipeline consisting of four core modules: Perception, Prediction, Path planning, and Control. The advent of multi-modal networks in recent years for several core modules has seen a convergence of sensors of varying modalities into a single unit. In such cases, there is a demand for data traffic up to ~2 GB/s, necessitating various hardware components \cite{SelfDrivingModule}. To facilitate this, several chip-based accelerators have been introduced, notably GPUs, NPUs, or deep learning accelerators provided by specific companies \cite{Lin2018TheAI}. 
\newline \hspace*{1em} In autonomous driving systems, four core modules play a critical role in the system's overall operation. First, Perception module \cite{li2022bevformer, lang2019pointpillars, liu2023bevfusion, hwang2022cramnet, zhou2018voxelnet} interprets sensor data from cameras, LiDAR, and radar employing to identify and classify objects like obstacles, road signs, and other vehicles, enhancing understanding of the vehicle's environment. Second, Prediction module \cite{liu2020driving, shao2023self, hubmann2017decision, geng2017scenario} which anticipates the actions of other objects like pedestrians, cyclists, and vehicles, can make informed decisions and ensure safe navigation. Third, Path planning module \cite{yuan2023evolutionary, huang2023differentiable, hu2023planning, yoon2022trajectory} determines the optimal path and behavior for the vehicle considering factors like road conditions, traffic laws, and the predicted actions of other road users. Finally, Control module \cite{williams2018information, kong2015kinematic} translates the planned trajectory into actionable commands including steering, acceleration, and braking for the vehicle's actuators.  
\newline \hspace*{1em}According to our research, the paper by D3 \cite{10.1145/3492321.3519576} was the first to apply an AI system architecture to autonomous driving systems. While there have been several proposals for communication and core modules platforms based on Service Level Objectives \cite{Casini2019ResponseTimeAO, quigley2009ros}, there has been a growing interest in comprehensive system frameworks capable of handling diverse pipelines, given the emergence of large-scale models and services that need to process a substantial number of queries. In autonomous driving systems, the necessity for a dynamic system capable of robust and safe operation across varying driving environments is paramount. For instance, in congested urban settings, intersections, or on highways, the inputs to the system vary significantly, necessitating a system that can adapt its core modules, including deep learning models, to these changing conditions through dynamic batching. While the D3 \cite{10.1145/3492321.3519576} addresses this need by implementing ERDOS, there is a notable lack of research in implementing such dynamic capabilities within the widely-used ROS framework. 
Thus, we aim to build upon the research conducted by \cite{10.1145/3492321.3519576}, further extending it within the ROS framework. 

\subsection{Hybrid Scheduler}
In the rapidly evolving field of computational scheduling, hybrid schedulers have become a key innovation for efficiently managing heterogeneous resources like CPUs and GPUs. Hybrid schedulers can be broadly classified into two categories. Firstly, hybrid schedulers designed for integrated computing resource management have concentrated on optimizing the coordination and utilization of heterogeneous resources such as CPUs and GPUs. Hybridhadoop \cite{oh2021hybridhadoop} dynamically manages CPU and GPU resources in response to real-time workload demands within Hadoop environments, an open-source software framework. Kleio \cite{doudali2019kleio} presents an advanced hybrid memory page scheduler that combines history-based data tiering with deep neural networks for optimized memory management, enhancing efficiency and enabling AI-driven resource allocation in computing systems.\cite{kaleem2014adaptive} presents two innovative adaptive scheduling algorithms for integrated CPU-GPU processors, enhancing heterogeneous computing by optimally utilizing both processor types for complex tasks, thus boosting system performance and efficiency. Second, Hybrid scheduling emphased fine-tuned adjustment and adaptability for specialized environments, facilitates tailored allocation of CPU and GPU memory to suit diverse scenarios. \cite{vasiliu2017hybrid} presents a hybrid scheduling algorithm designed for big data environments, effectively managing complex and large-scale operations while balancing efficiency and robust task processing. The schedGPU \cite{reano2016schedgpu}, a fine-grained and adaptive scheduler, marks a major advancement in GPU resource management by dynamically optimizing GPU utilization for intensive tasks and adjusting to workload changes to enhance performance.

\section{NeuroFlow}

\begin{figure*}[ht]
    \centering
    \includegraphics[width=\textwidth]{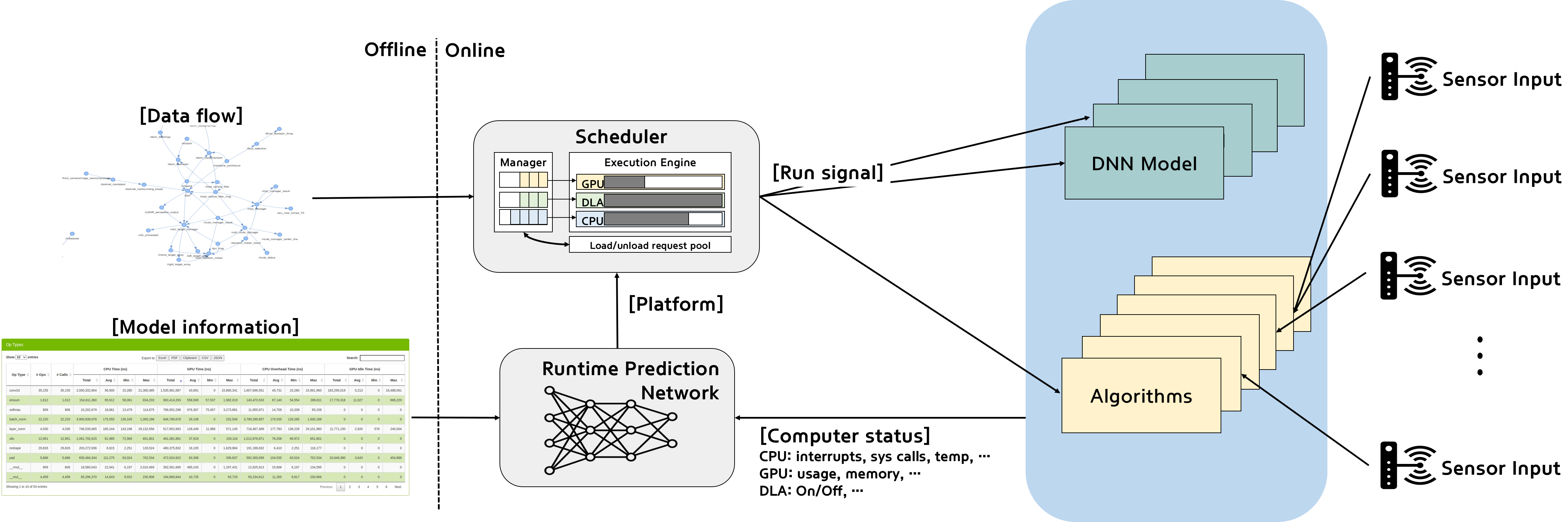}
    \caption{Overview of proposed system, NeuroFlow}
    \label{fig1:overview}
\end{figure*}

\subsection{Overview of proposed system}
NeuroFlow is a system composed of a runtime prediction for inferring the execution platform and a hybrid scheduler to enhance parallelism by integrating information from computing resources and pre-trained deep learning models. By categorizing algorithms that utilize deep learning models and CPU-intensive algorithms, the proposed system efficiently allocates platform resources, avoiding wastage. A overview of the proposed system is presented in Figure \ref{fig1:overview}.

To achieve this, the system divides tasks into parallel queues and proceeds with resource allocation. Initially, two separate queues are created, classifying all programs, and each queue is designed to be processed as parallelly as possible. In flows requiring deep learning, the system determines the platform through runtime prediction, while all other programs are scheduled based on the priority of the data flow. A simplified pseudocode for the system is presented in Algorithm \ref{algorithm: pseudocode}.

\begin{algorithm}
    \caption{Pseudocode of the NeuroFlow Scheduling System}
    \label{algorithm: pseudocode}
    \begin{algorithmic}[1]
        \State Initialize DNN queue, Non-DNN queue
        \State Process programs of DNN queue and programs in Non-DNN queue in parallel
        \For{each program $p$}
            \State $computer\_status \gets$ get\_computer\_status()
            \If{program\_uses\_DNN($p$)}
                \State Add $p$ to DNN queue
                \State $predicted\_platform \gets$ \\ \hspace{\algorithmicindent}\hspace{\algorithmicindent}\hspace{\algorithmicindent}predict\_runtime($p$, $computer\_status$)
                \State Execute $p$ on $predicted\_platform$
            \Else
                \State Add $p$ to Non-DNN queue
                \If{is\_DAG($p$)}
                    \State Set higher priority for $p$
                \EndIf
                \State Schedule $p$ using CFS
            \EndIf
        \EndFor
    \end{algorithmic} 
\end{algorithm}

\subsection{Hybrid scheduler for parallelism}
The hybrid scheduler designed in the previous research focused on adjusting and optimizing the utilization of resources across various platforms. Similarly, the proposed scheduler also emphasizes the optimization of resource utilization and system stability by leveraging platform predictions through pre-defined data and a runtime prediction model.
\subsubsection{Pre-define Data flow graph}
\label{subscection graph}
Having executed all the programs that comprise the autonomous driving system, we conducted an a priori analysis of the data flow, assuming that no nodes initiate new subscriptions or publications during program execution. By depicting the data flow in a graphical format, we enable the analysis of data dependencies within each program. This understanding, taking into account the system's inherent interconnections of perception, localization, decision-making, and control, provides a foundation for system design that accommodates its synchronous and asynchronous characteristics.
The overarching system structure exhibits a graph with a non-cyclic, or acyclic, structure due to the feedback nature of control components. In an effort to capture the system's synchronous characteristics, we aimed to identify Directed Acyclic Graph (DAG) subgraphs within this structure. We identified end nodes with an out-degree of zero and their ancestor nodes, creating subgraphs, and subsequently, determined the subgraphs using a topological sorting algorithm.
Within DAG subgraphs, we predict independent flows, allowing for parallel execution, either across different platforms or within the same platform, while dynamically allocating resources as needed. Nodes within the subgraph are assigned priority weights in a primary consideration of the topological ordering.

\subsubsection{Efficient Resource Allocation}
\begin{figure}[ht]
    \centering
    \includegraphics[width=\columnwidth]{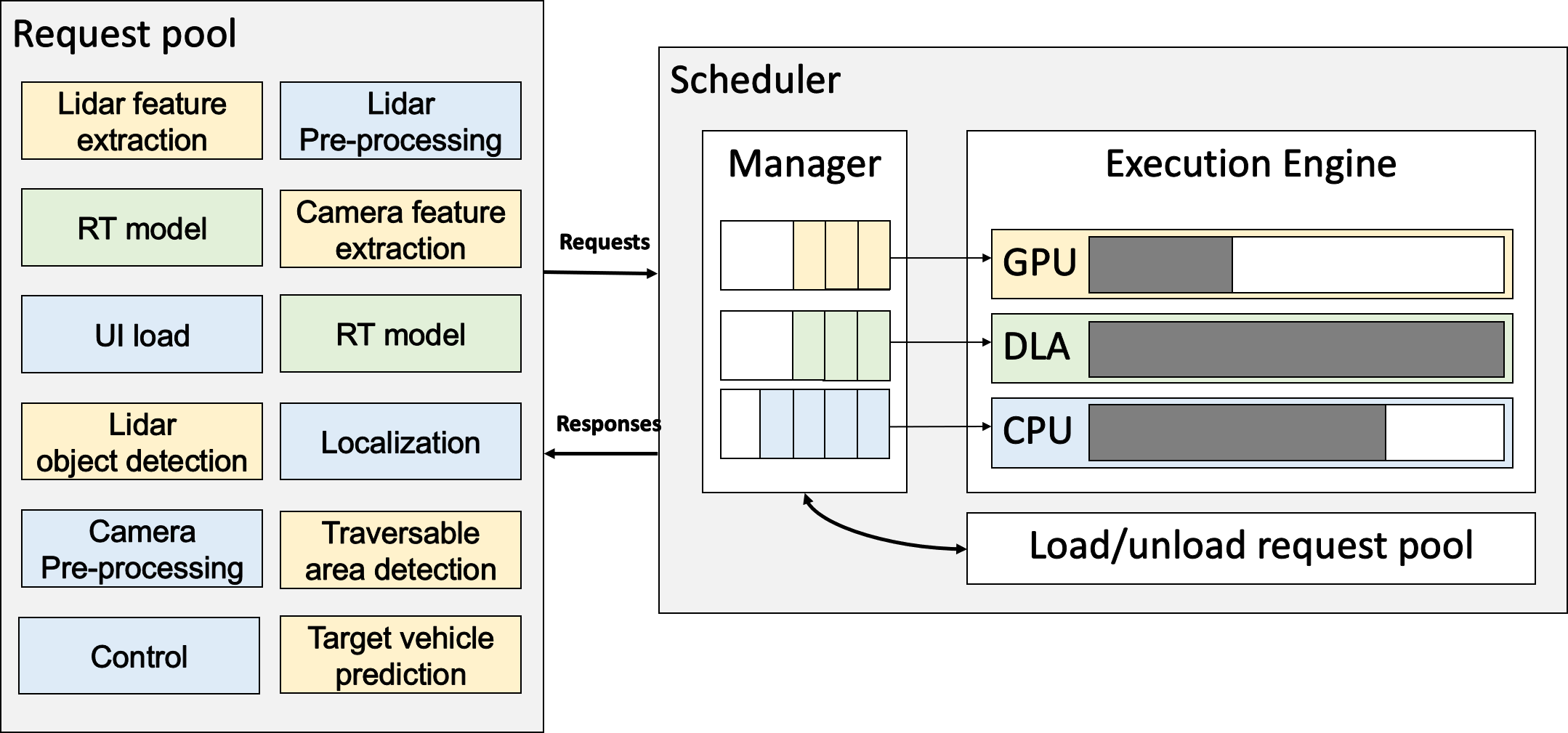}
    \caption{Hybrid scheduler of the proposed system}
    \label{fig1:scheduelr}
\end{figure}

This system adopts a Primary-Secondary structure, where the scheduler, equipped with dual queues, performs control functions for the program. As depicted in Figure \ref{fig1:scheduelr}, various programs within the autonomous driving system are classified based on their resource requirements. The manager oversees the queues, and the scheduler executes programs according to two policies, aiming to efficiently utilize resources and maintain stability through parallelism considerations. Firstly, it incorporates the Completely Fair Scheduler(CFS) to ensure resource fairness. This guarantees equitable resource allocation among DAG-shaped subgraphs of DAG(\ref{subscection graph}) by utilizing Nice values to determine scheduling priorities. This approach ensures that subgraphs with higher priorities receive more resources, providing fair resource allocation even to subgraphs with lower weights. Secondly, it employs priority-based scheduling within DAG-shaped subgraphs. Priorities are assigned among nodes based on a predefined data flow, with higher priorities given to nodes farther away from end nodes. This strategy facilitates the efficient distribution of resources. Such a system enables efficient scheduling in distributed autonomous driving systems, providing fair resource distribution and stability.

\begin{figure}[ht]
    \centering
    \caption{Detector model runtime distribution on dynamic batch}
    \includegraphics[width=0.5\textwidth]{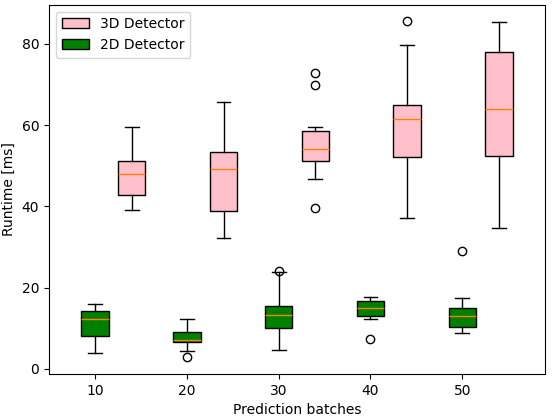}
    \label{fig:dynamic}
\end{figure}

\section{Platform Runtime Estimation}
This section describes the methodology for developing a predictor of runtime operational performance for each model, specific to different platforms. It also includes an explanation of the prior knowledge required for this approach. Additionally, the section emphasizes the necessity of employing a simple Attention-based model to ensure rapid inference performance.
\subsection{Computing Resources Required Solely for Inference}
In fields where cutting-edge technology and practical implementation, such as autonomous driving, intertwine, there are instances where technology is deployed before it is fully optimized. Consequently, real-world operations can lead to unforeseen events, differing from initial predictions. This is largely because the CPU must process the data entering the model and handle the communication bus with the GPU. The role of the CPU extends beyond mere data processing; it significantly impacts the overall efficiency and performance of the system. Regrettably, issues such as CPU exhaustion become more pronounced in situations where the surrounding environment grows complex, necessitating more sophisticated control and sensing. This puts a strain on the computer's resources. While it varies depending on the model's characteristics, models with tracking features show a noticeable increase in input time as the input batch size increases. This trend is clearly visible in Fig \ref{fig:dynamic}, where the escalation in input time is distinctly evident. Therefore, effective management of CPU resources is essential for optimizing model performance.

While there has been considerable focus on GPUs, our comprehensive approach considers the entire computer system, including the Ethernet bus, CPU, GPU, and other peripheral buses. This holistic investigation reveals several key observations: 1) The nature of the data source inputs is crucial. 2) There is a significant correlation between detailed metrics such as CPU iowait and context switches. 3) The size of the model's output impacts not just the model's throughput but also the overall performance of its inference algorithm. These insights emphasize the importance of viewing system resources in their entirety, beyond the traditional focus on model operations or GPU performance alone.

\subsection{Anatomy of Deep Learning Models}
Deep learning models, consisting of key modules like Convolution, BatchNorm, Activation, and recent additions like Attention mechanisms, exhibit varying characteristics. A crucial aspect to understand is that not all modules, specifically kernel operations, incur the same latency. This variance in delay is not consistent across different hardware platforms such as GPUs, CPUs, and DLAs.

Furthermore, the field of autonomous driving perception has been evolving with the adoption of structures like Large Kernel CNNs \cite{ding2022scaling, vasu2023fastvit} or Transformers \cite{dosovitskiy2020image}to achieve wider Receptive Fields. However, this advancement comes at the cost of significant overhead, manifesting in increased GPU usage and larger memory requirements. Despite these challenges, our approach prioritizes the FLOPs of Conv and Linear layers, along with the total FLOPs of the model, as the primary features for predicting the runtime of the model (see model's detail in Table \ref{tab:comparison}. This strategy aims to balance the need for advanced model capabilities with the practical constraints of computational resources.

\subsection{Model Inference Runtime Prediction}
The field of model inference time prediction has been gaining significant attention recently, especially in the realms of efficient AI systems and deep learning model architecture search. Rapid advancements and innovative approaches are emerging in this area. Among these, there are kernel-level and model-level predictions, focusing on estimating the required computational speed of the model itself. This involves calculating the required number of Floating Operations or using MACs (Multiply-Accumulates) for baseline recognizers, a trend that is increasingly being adopted as evidenced by studies like Zhang et al. 2021 \cite{zhang2021nn}.

In our work, we aimed to develop an estimator that is more suited to the dynamic nature of operational loads in real-world driving scenarios, where computational demands can actively fluctuate based on input batches or typical driving conditions. Our approach not only considers the computational requirements of key model modules such as \texttt{Convolution (Conv)}, \texttt{Batch Normalization (BN)}, \texttt{Activation}, and Large Size \texttt{Matrix Multiplication (MatMul)} but also gathers state information of the platform's CPU, GPU, and memory. By utilizing this data as input, we developed a model that predicts the execution time of a given model on various platforms. This approach provides a more comprehensive understanding of model performance in real-world applications, considering both the model's intrinsic computational needs and the state of the underlying hardware platform. 
As a result of our approach, we have achieved higher performance compared to existing kernel-level studies, as demonstrated in our findings (see Table \ref{tab:approchComparison}).

\subsubsection{Model Design}
Our NeuroFlow's real-time system employs a Runtime Prediction model for scheduling policies, as revealed in previous research \cite{zhang2021nn}. This model, based on an MLP (Multi-Layer Perceptron) network, is capable of rapid inference even in CPU systems. To understand the correlations among different domains of accelerators like GPUs, CPUs, and DLAs, we utilized an Attention-based Token Mixer. This approach allowed us to use the model-level information of the deep learning model being predicted as a query to perform attention on the states of GPUs and CPUs. The data was trained by applying various loads on different platforms and deep learning models, including diverse CPU and GPU loads.
Additionally, the model outputs logits representing the potential platforms on which the given model can perform inference. This process involves passing the output through an MLP classifier after the token mixer stage and training it using Cross-Entropy (CE) loss. Such an approach enhances the robustness of the model and aids in learning appropriate tendencies from multi-platform data, ensuring its effectiveness across various hardware environments.
The model, structured in this manner, guarantees a parameter size of 12KB and an inference speed of 600μs when running on a GPU.

\begin{table}[]
    \small
    \caption{Platform level estimator achieves higher accuracy. test with 4 models}
    \label{tab:approchComparison}
    \begin{tabular}{c|cc|cc}
    \toprule
    \multicolumn{1}{c}{\multirow{2}{*}{Device}} & \multicolumn{2}{|c|}{Kernel Level \cite{zhang2021nn}} & \multicolumn{2}{c}{Platform Level (ours)} \\ 
    \multicolumn{1}{c|}{} & RMSE              & +10\% Acc.         & RMSE              & +10\% Acc.         \\
    \midrule
    IPC GPU & 15.32 ms & 79.2\% & 4.5 ms & 97.3\% \\
    Jetson GPU & 140.3 ms & 20.4\% & 23.4 ms & 9\%\\
    Jetson DLA & 96.3 ms & 34.1\% & 9.4 ms & 92\%  \\
    \bottomrule
    \end{tabular}
\end{table}

\section{Evaluation}
We evaluated NeuroFlow in a real-world setup, consisting of a Self-Driving Vehicle equipped with one Industrial PC and an SBC (in Table \ref{tab:systemsetup}, Fig \ref{fig:configuration}). Additionally, the evaluation was conducted using a test set
\subsection{Experiment Setup}
The runtime prediction proposed in this study is a post-training prediction, differing from the prior predictions used in NAS (Neural Architecture Search) or network design fields. Consequently, there are no publicly available benchmarks, and it is challenging to obtain fair data, as most are based on specialized hardware such as NPUs (Neural Processing Units) and VPUs (Vision Processing Units). 
\subsubsection{Comparison baseline}
The system we propose is a scheduling software designed for autonomous driving systems, where it's crucial to consider metrics that reflect the overall system's fairness. The stability and response time of the final output commands (vehicle control inputs) are key indicators of the system's reliability. We compare our proposed system configuration with a conventional system currently in use by urban autonomous driving buses. (See Table \ref{tab:systemsetup}.) This comparison helps us to gauge the effectiveness of our approach. Additionally, for evaluating the inference time estimator, we conducted a comparative analysis of the model's performance using FLOPs and a combined FLOPs+MACs kernel-level predictor.

\subsubsection{Metrics}

\begin{table*}[htp]
\caption{Runtime Estimator Result for latency on Desktop GPU, SBC GPU, DLA}
\label{tab:estimatorResult}
\centering
\begin{tabular}{c|ccccc}
\hline 
Model variants & RMSE               & RMSPE             & $\pm5\%$ Acc. & $\pm10\%$ Acc.       & Cls. Acc.      \\ \hline
IPC GPU & 5.83 ms  & 38.45 ms & 80.85\% & 97.17\% & 89.57\% \\
Jetson GPU & 50.38 ms & 43.02 ms & 38.6\% & 53.1\% & 45.5\% \\
Jeton DLA  & 3.86 ms & 15.26 ms & 93.59\% & 93.04\% & 94.1\% \\
\hline
\end{tabular}
\end{table*}

To evaluate the classification and regression performance of these models, we employed metrics such as Root Mean Square Error (RMSE) and Root Mean Square Percentage Error (RMSPE), along with accuracy rates for classification tasks. Moreover, due to the fluctuations in model runtime predictions and the very small time units involved, we utilized accuracy rates with margins of $\pm5\%$ and $\pm10\%$, as suggested by Zela et al. 2019 \cite{zela2019bench}, to assess the predictions' accuracy.
\subsection{Platform Runtime Prediction Evaluation}
For comparison purposes, we established an environment capable of autonomously running deep learning models on various platforms and collecting stats for about 24 hours of CPU system operation. We also configured a system where IPCs and SBCs are interconnected, allowing them to execute their respective models and exchange system status information. This setup is particularly useful for later utilization in schedulers. Furthermore, we simulated and collected data for all possible combinations of independent and concurrent executions of the deep learning models (in Table \ref{tab:comparison}). The collected data is consolidated into a single dataset, which includes time-filtered records. These records also account for the runtime prediction model and a one-second lag to compensate for communication overhead, treating the prediction performance after this delay as the correct output.

\begin{figure}[ht]
    \centering
    \includegraphics[width=\columnwidth]{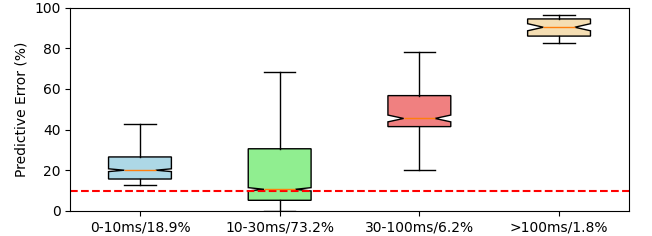}
    \caption{Prediction Errors in Interval-Divided Regions. \\ the red dashed line is drawn at 10\%}
    \label{fig:latencyError}
\end{figure}

\begin{figure}[ht]
    \centering
    \includegraphics[width=0.8\columnwidth]{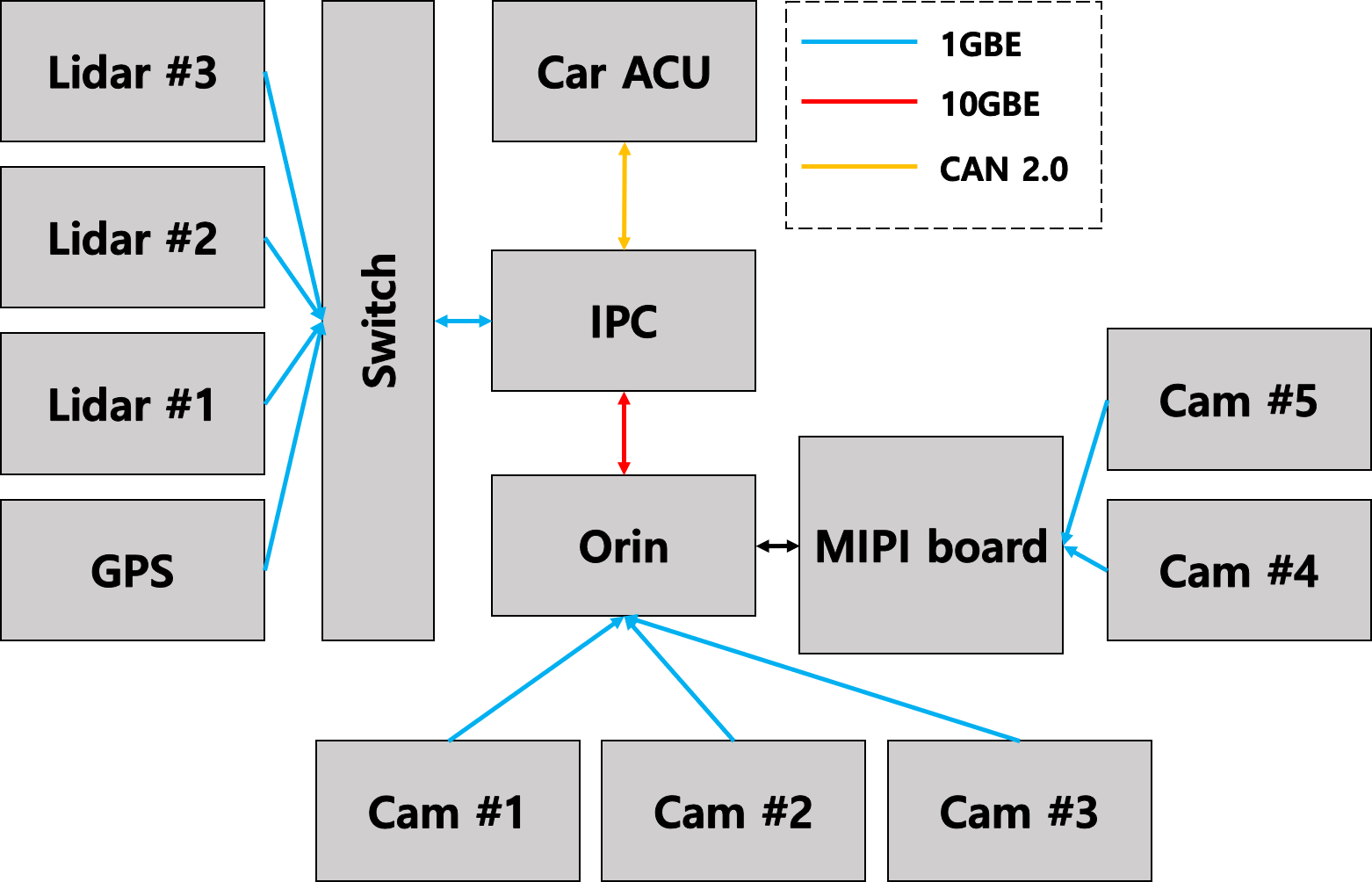}
    \caption{System configuration}
    \label{fig:configuration}
\end{figure}

The results of this data are organized in a results table (in Table \ref{tab:estimatorResult}), detailing the runtime prediction performance and classifier performance for each platform. Due to the minuscule millisecond units involved in model execution times, we examined the model's performance using error distribution intervals for 10\% Error (see  \ref{fig:latencyError}). We observed that models with relatively faster execution times exhibited higher performance. This is particularly noteworthy as achieving accuracy with very small values can be challenging; even minor data variations can lead to significant errors. Conversely, models with longer execution times might require additional data for accuracy. In some cases, models with tracking features, when devoid of input batches, inferred at high speeds, acting as outliers and consequently showing lower accuracy rates.

\begin{table}[]
    \small
    \caption{Comparison System Specification and Setup}
    \label{tab:systemsetup}
    \begin{tabularx}{0.5\textwidth}{cc|c}
        
        \hline
        \multirow{2}{*}{System setups} & \multicolumn{2}{c}{Specification} \\
                                       & IPC             & SBC             \\ \hline
        \multicolumn{1}{c|}{\multirow{2}{*}{Components}} &
          \begin{tabular}[c]{@{}c@{}}Intel E2176\\ 32GB of RAM\end{tabular} &
          \multicolumn{1}{c}{ARM A78AE} \\ \cline{2-3} 
        \multicolumn{1}{c|}{} &
          \begin{tabular}[c]{@{}c@{}}NV TU104-400A\\ FP32 : 10.07TFLOPs\end{tabular} &
          \multicolumn{1}{c}{\begin{tabular}[c]{@{}c@{}}NVIDIA GA10B\\ FP32: 5.3TFLOPS\\ INT8:275 TOPS (DLA)\end{tabular}} \\ \hline
        Conventional                   & 3 IPC           & 1 SBC           \\ \hline
        Ours                           & 1 IPC           & 1 SBC           \\ \hline
    \end{tabularx}
\end{table}

\begin{table*}
  \centering
  \caption{The information for each model represents factors}
  \label{tab:comparison}
  \begin{threeparttable}
  \begin{tabular}{c|cccc}
    \toprule
     &3D obejct detection&2D object detection&Trajectory prediction& Traversable area detection \\
    \midrule
    Weight file size & 10.5 MB& 75.6MB & 323.3MB & 13.4 MB \\
    Trainable \# of param & 4.86 M & 36.91 M & 3.78 M & 1.08 M \\
    Non-trainable \# of param & 0 & 0 & 24.45 M & 0\\
    Total \# of param & 4.86 M &  36.91 M & 28.23 M &  1.08 M\\
    \# of conv layer & 19 & 92 & 62 & 87 \\
    \# of linear layer & 1 & 0 & 28 & 12 \\
    FLOPs & 250.4G & 52.257G & 130.3G & 18.67G \\
    Framework & Torch, TensorRT & Torch & Torch & Torch, TensorRT\\
    Baseline & \cite{kim2023lidar} & \cite{wang2023yolov7} & \cite{kim2023non} & -\\
    \midrule
    RAM\tnote{*} & 2763MiB/5851MiB & 2311/2501MiB & 2791MiB & 1707MiB/3227MiB \\
    Avg inference time\tnote{*} & 50.4 ms & 11.1 ms & 31 ms & 8.5 ms\\
  \bottomrule
\end{tabular}
\begin{tablenotes}       
    \item[*] This information pertains to each model when executed individually.
\end{tablenotes}
\end{threeparttable}
\end{table*}

{\small
\bibliographystyle{ieee_fullname}
\bibliography{egbib}

\begin{thebibliography}{10}\itemsep=-1pt

\bibitem{apolloplatform}
{ApolloAuto}.
\newblock {Apollo Platform Repository}.
\newblock \url{https://github.com/ApolloAuto/apollo-platform/tree/master}.

\bibitem{autoware}
{Autoware}.
\newblock {Autoware User's Manual - Document Version 1.1}.
\newblock \url{https://tinyurl.com/2v2jkk9n}.

\bibitem{Caradas}
{Caradas}.
\newblock Sae autonomous level 3: Levels of driving automation.
\newblock \url{https://caradas.com/sae-autonomous-level-3-levels-of-driving-automation/}.
\newblock Accessed: October 23, 2023.

\bibitem{Casini2019ResponseTimeAO}
Daniel Casini, Tobias Blass, Ingo L{\"u}tkebohle, and Bj{\"o}rn~B. Brandenburg.
\newblock Response-time analysis of ros 2 processing chains under reservation-based scheduling.
\newblock In {\em Euromicro Conference on Real-Time Systems}, 2019.

\bibitem{ding2022scaling}
Xiaohan Ding, Xiangyu Zhang, Jungong Han, and Guiguang Ding.
\newblock Scaling up your kernels to 31x31: Revisiting large kernel design in cnns.
\newblock In {\em Proceedings of the IEEE/CVF conference on computer vision and pattern recognition}, pages 11963--11975, 2022.

\bibitem{dosovitskiy2020image}
Alexey Dosovitskiy, Lucas Beyer, Alexander Kolesnikov, Dirk Weissenborn, Xiaohua Zhai, Thomas Unterthiner, Mostafa Dehghani, Matthias Minderer, Georg Heigold, Sylvain Gelly, et~al.
\newblock An image is worth 16x16 words: Transformers for image recognition at scale.
\newblock {\em arXiv preprint arXiv:2010.11929}, 2020.

\bibitem{doudali2019kleio}
Thaleia~Dimitra Doudali, Sergey Blagodurov, Abhinav Vishnu, Sudhanva Gurumurthi, and Ada Gavrilovska.
\newblock Kleio: A hybrid memory page scheduler with machine intelligence.
\newblock In {\em Proceedings of the 28th International Symposium on High-Performance Parallel and Distributed Computing}, pages 37--48, 2019.

\bibitem{geng2017scenario}
Xinli Geng, Huawei Liang, Biao Yu, Pan Zhao, Liuwei He, and Rulin Huang.
\newblock A scenario-adaptive driving behavior prediction approach to urban autonomous driving.
\newblock {\em Applied Sciences}, 7(4):426, 2017.

\bibitem{10.1145/3492321.3519576}
Ionel Gog, Sukrit Kalra, Peter Schafhalter, Joseph~E. Gonzalez, and Ion Stoica.
\newblock D3: A dynamic deadline-driven approach for building autonomous vehicles.
\newblock In {\em Proceedings of the Seventeenth European Conference on Computer Systems}, EuroSys '22, page 453–471, New York, NY, USA, 2022. Association for Computing Machinery.

\bibitem{hu2023planning}
Yihan Hu, Jiazhi Yang, Li Chen, Keyu Li, Chonghao Sima, Xizhou Zhu, Siqi Chai, Senyao Du, Tianwei Lin, Wenhai Wang, et~al.
\newblock Planning-oriented autonomous driving.
\newblock In {\em Proceedings of the IEEE/CVF Conference on Computer Vision and Pattern Recognition}, pages 17853--17862, 2023.

\bibitem{huang2023differentiable}
Zhiyu Huang, Haochen Liu, Jingda Wu, and Chen Lv.
\newblock Differentiable integrated motion prediction and planning with learnable cost function for autonomous driving.
\newblock {\em IEEE transactions on neural networks and learning systems}, 2023.

\bibitem{hubmann2017decision}
Constantin Hubmann, Marvin Becker, Daniel Althoff, David Lenz, and Christoph Stiller.
\newblock Decision making for autonomous driving considering interaction and uncertain prediction of surrounding vehicles.
\newblock In {\em 2017 IEEE intelligent vehicles symposium (IV)}, pages 1671--1678. IEEE, 2017.

\bibitem{hwang2022cramnet}
Jyh-Jing Hwang, Henrik Kretzschmar, Joshua Manela, Sean Rafferty, Nicholas Armstrong-Crews, Tiffany Chen, and Dragomir Anguelov.
\newblock Cramnet: Camera-radar fusion with ray-constrained cross-attention for robust 3d object detection.
\newblock In {\em European Conference on Computer Vision}, pages 388--405. Springer, 2022.

\bibitem{kaleem2014adaptive}
Rashid Kaleem, Rajkishore Barik, Tatiana Shpeisman, Brian~T Lewis, Chunling Hu, and Keshav Pingali.
\newblock Adaptive heterogeneous scheduling for integrated gpus.
\newblock In {\em Proceedings of the 23rd international conference on Parallel architectures and compilation}, pages 151--162, 2014.

\bibitem{kim2023lidar}
Jongho Kim and Kyongsu Yi.
\newblock Lidar object perception framework for urban autonomous driving: detection and state tracking based on convolutional gated recurrent unit and statistical approach.
\newblock {\em IEEE Vehicular Technology Magazine}, 2023.

\bibitem{kim2023non}
Yujin Kim, Eunbin Seo, Chiyun Noh, and Kyoungsu Yi.
\newblock Non-autoregressive transformer based ego-motion independent pedestrian trajectory prediction on egocentric view.
\newblock {\em IEEE Access}, 2023.

\bibitem{kong2015kinematic}
Jason Kong, Mark Pfeiffer, Georg Schildbach, and Francesco Borrelli.
\newblock Kinematic and dynamic vehicle models for autonomous driving control design.
\newblock In {\em 2015 IEEE intelligent vehicles symposium (IV)}, pages 1094--1099. IEEE, 2015.

\bibitem{lang2019pointpillars}
Alex~H Lang, Sourabh Vora, Holger Caesar, Lubing Zhou, Jiong Yang, and Oscar Beijbom.
\newblock Pointpillars: Fast encoders for object detection from point clouds.
\newblock In {\em Proceedings of the IEEE/CVF conference on computer vision and pattern recognition}, pages 12697--12705, 2019.

\bibitem{li2022bevformer}
Zhiqi Li, Wenhai Wang, Hongyang Li, Enze Xie, Chonghao Sima, Tong Lu, Yu Qiao, and Jifeng Dai.
\newblock Bevformer: Learning bird’s-eye-view representation from multi-camera images via spatiotemporal transformers.
\newblock In {\em European conference on computer vision}, pages 1--18. Springer, 2022.

\bibitem{Lin2018TheAI}
Shi-Chieh Lin, Yunqi Zhang, Chang-Hong Hsu, Matt Skach, Md~E. Haque, Lingjia Tang, and Jason Mars.
\newblock The architectural implications of autonomous driving: Constraints and acceleration.
\newblock {\em Proceedings of the Twenty-Third International Conference on Architectural Support for Programming Languages and Operating Systems}, 2018.

\bibitem{liu2020driving}
Shiwen Liu, Kan Zheng, Long Zhao, and Pingzhi Fan.
\newblock A driving intention prediction method based on hidden markov model for autonomous driving.
\newblock {\em Computer Communications}, 157:143--149, 2020.

\bibitem{liu2023bevfusion}
Zhijian Liu, Haotian Tang, Alexander Amini, Xinyu Yang, Huizi Mao, Daniela~L Rus, and Song Han.
\newblock Bevfusion: Multi-task multi-sensor fusion with unified bird's-eye view representation.
\newblock In {\em 2023 IEEE international conference on robotics and automation (ICRA)}, pages 2774--2781. IEEE, 2023.

\bibitem{SelfDrivingModule}
{Nicolo Valigi}.
\newblock {Lessons Learned Building a Self-Driving Car on ROS}.
\newblock \url{https://roscon.ros.org/2018/presentations/ROSCon2018_ LessonsLearnedSelfDriving.pdf}, 2018.

\bibitem{oh2021hybridhadoop}
Chanyoung Oh, Hyeonjin Jung, Saehanseul Yi, Illo Yoon, and Youngmin Yi.
\newblock Hybridhadoop: Cpu-gpu hybrid scheduling in hadoop.
\newblock In {\em The International Conference on High Performance Computing in Asia-Pacific Region}, pages 40--49, 2021.

\bibitem{quigley2009ros}
Morgan Quigley, Ken Conley, Brian Gerkey, Josh Faust, Tully Foote, Jeremy Leibs, Rob Wheeler, and Andrew~Y Ng.
\newblock {ROS: An OpenSource Robot Operating System}.
\newblock In {\em Proceedings of the IEEE International Conference on Robotics and Automation (ICRA); Workshop on Open Source Robotics}, volume~3, page~5, May 2009.

\bibitem{reano2016schedgpu}
Carlos Reano, Federico Silla, and Matthew~J Leslie.
\newblock schedgpu: fine-grain dynamic and adaptative scheduling for gpus.
\newblock In {\em 2016 International Conference on High Performance Computing \& Simulation (HPCS)}, pages 993--997. IEEE, 2016.

\bibitem{SAEJ3016}
{SAE International}.
\newblock Taxonomy and definitions for terms related to on-road motor vehicle automated driving systems.
\newblock \url{https://www.sae.org/standards/content/j3016_201806/}, 2014.
\newblock SAE Standard J3016.

\bibitem{shao2023self}
Wenbo Shao, Jun Li, and Hong Wang.
\newblock Self-aware trajectory prediction for safe autonomous driving.
\newblock {\em arXiv preprint arXiv:2305.09147}, 2023.

\bibitem{vasiliu2017hybrid}
Laura Vasiliu, Florin Pop, Catalin Negru, Mariana Mocanu, Valentin Cristea, and Joanna Kolodziej.
\newblock A hybrid scheduler for many task computing in big data systems.
\newblock {\em International Journal of Applied Mathematics and Computer Science}, 27(2):385--399, 2017.

\bibitem{vasu2023fastvit}
Pavan Kumar~Anasosalu Vasu, James Gabriel, Jeff Zhu, Oncel Tuzel, and Anurag Ranjan.
\newblock Fastvit: A fast hybrid vision transformer using structural reparameterization.
\newblock {\em arXiv preprint arXiv:2303.14189}, 2023.

\bibitem{wang2023yolov7}
Chien-Yao Wang, Alexey Bochkovskiy, and Hong-Yuan~Mark Liao.
\newblock Yolov7: Trainable bag-of-freebies sets new state-of-the-art for real-time object detectors.
\newblock In {\em Proceedings of the IEEE/CVF Conference on Computer Vision and Pattern Recognition}, pages 7464--7475, 2023.

\bibitem{williams2018information}
Grady Williams, Paul Drews, Brian Goldfain, James~M Rehg, and Evangelos~A Theodorou.
\newblock Information-theoretic model predictive control: Theory and applications to autonomous driving.
\newblock {\em IEEE Transactions on Robotics}, 34(6):1603--1622, 2018.

\bibitem{yoon2022trajectory}
Youngmin Yoon and Kyongsu Yi.
\newblock Trajectory prediction using graph-based deep learning for longitudinal control of autonomous vehicles: A proactive approach for autonomous driving in urban dynamic traffic environments.
\newblock {\em IEEE Vehicular Technology Magazine}, 17(4):18--27, 2022.

\bibitem{yuan2023evolutionary}
Kang Yuan, Yanjun Huang, Shuo Yang, Zewei Zhou, Yulei Wang, Dongpu Cao, and Hong Chen.
\newblock Evolutionary decision-making and planning for autonomous driving based on safe and rational exploration and exploitation.
\newblock {\em Engineering}, 2023.

\bibitem{zela2019bench}
Arber Zela, Julien Siems, and Frank Hutter.
\newblock Nas-bench-1shot1: Benchmarking and dissecting one-shot neural architecture search.
\newblock In {\em International Conference on Learning Representations}, 2019.

\bibitem{zhang2021nn}
Li~Lyna Zhang, Shihao Han, Jianyu Wei, Ningxin Zheng, Ting Cao, Yuqing Yang, and Yunxin Liu.
\newblock Nn-meter: Towards accurate latency prediction of deep-learning model inference on diverse edge devices.
\newblock In {\em Proceedings of the 19th Annual International Conference on Mobile Systems, Applications, and Services}, pages 81--93, 2021.

\bibitem{zhou2018voxelnet}
Yin Zhou and Oncel Tuzel.
\newblock Voxelnet: End-to-end learning for point cloud based 3d object detection.
\newblock In {\em Proceedings of the IEEE conference on computer vision and pattern recognition}, pages 4490--4499, 2018.

\end{thebibliography}
}

\end{document}